\documentclass{article}

\usepackage[utf8]{inputenc} 
\usepackage[T1]{fontenc}    
\usepackage{hyperref}       
\usepackage{url}            
\usepackage{amsfonts}       
\usepackage{nicefrac}       
\usepackage{microtype}      
\usepackage{graphicx}
\usepackage{subcaption}
\usepackage{color}
\usepackage{verbatim}
\usepackage{booktabs}
\usepackage[square, comma, numbers]{natbib}
\usepackage[final]{nips_2018}

\title{Few-Shot Generalization Across Dialogue Tasks}

\author{
  Vladimir Vlasov, Akela Drissner-Schmid, Alan Nichol \\
  Rasa \\
  \texttt{<firstname>@rasa.com}
}

\begin{document}

\maketitle

\begin{abstract}
Machine-learning based dialogue managers are able to learn complex behaviors in order to complete a task, but it is not straightforward to extend their capabilities to new domains. We investigate different policies' ability to handle uncooperative user behavior, and how well expertise in completing one task (such as restaurant reservations) can be reapplied when learning a new one (e.g. booking a hotel). We introduce the Recurrent Embedding Dialogue Policy (REDP), which embeds system actions and dialogue states in the same vector space. REDP contains a memory component and attention mechanism based on a modified Neural Turing Machine, and significantly outperforms a baseline LSTM classifier on this task. We also show that both our architecture and baseline solve the bAbI dialogue task, achieving $100\%$ test accuracy.
\end{abstract}

\section{Introduction}

Conversational software is an area of increasing importance, with the growth of voice-controlled devices and applications built inside of messaging services. A grand challenge in this field is to create software which is capable of holding extended conversations, carrying out tasks, keeping track of conversation history, and coherently responding to new information. There is no hope of explicitly codifying every possible interaction a human user may have with a conversational system, and as such there is a large body of work on statistical dialogue systems, where machine learning is used in order to understand spoken or written language and to guide the conversation. 

A recurring theme in the literature on task-oriented dialogue systems is the necessity (or lack thereof) of the application developer providing domain-specific code or knowledge. At one end of the spectrum are fully hand-crafted systems, such as those built with AIML, in which the developer has to codify every possible conversation manually. At the other extreme are so-called `end-to-end' systems, which are trained on unannotated dialogue transcripts and do not include any domain-specific logic or code. In-between these extremes lie approaches which combine domain-specific code with probabilistic components trained using dialogue data. A well-known deficiency of such systems is that once trained, it is not straightforward to adapt these to new domains~\citep{mrkvsic2015multi, gavsic2014incremental, williams2013multi}. If a dialogue system requires less task-specific code, it should be easy to increase the number of tasks it can complete, assuming that training data is available for each. For many developers, collecting a large number of conversations as training data is a prohibitive barrier, so we are especially interested in learning from small datasets.

\subsection{Cross-task Generalization in Dialogue}\label{sec:generalisation}

Slot filling involves identifying a user's goal and then collecting the information required to complete that task. For example, a restaurant search may require a cuisine, location, and price range before making a recommendation. These fields are typically called slots. We define \emph{cooperative} dialogue to be the case where the user responds to system prompts with the requested information. It is easy for a developer to implement these cooperative dialogues, for example by using a skeletal dialogue manager consisting of a single \texttt{while} loop which asks for missing information until all required slots are filled. We define \emph{uncooperative} dialogue as any deviation from this. It is infeasible to handle all possible deviations explicitly, so we are primarily interested in enabling developers to build software that can handle the complexity of uncooperative dialogue through the help of generalization.

\begin{figure}
\centering
\includegraphics[width=0.6\linewidth]{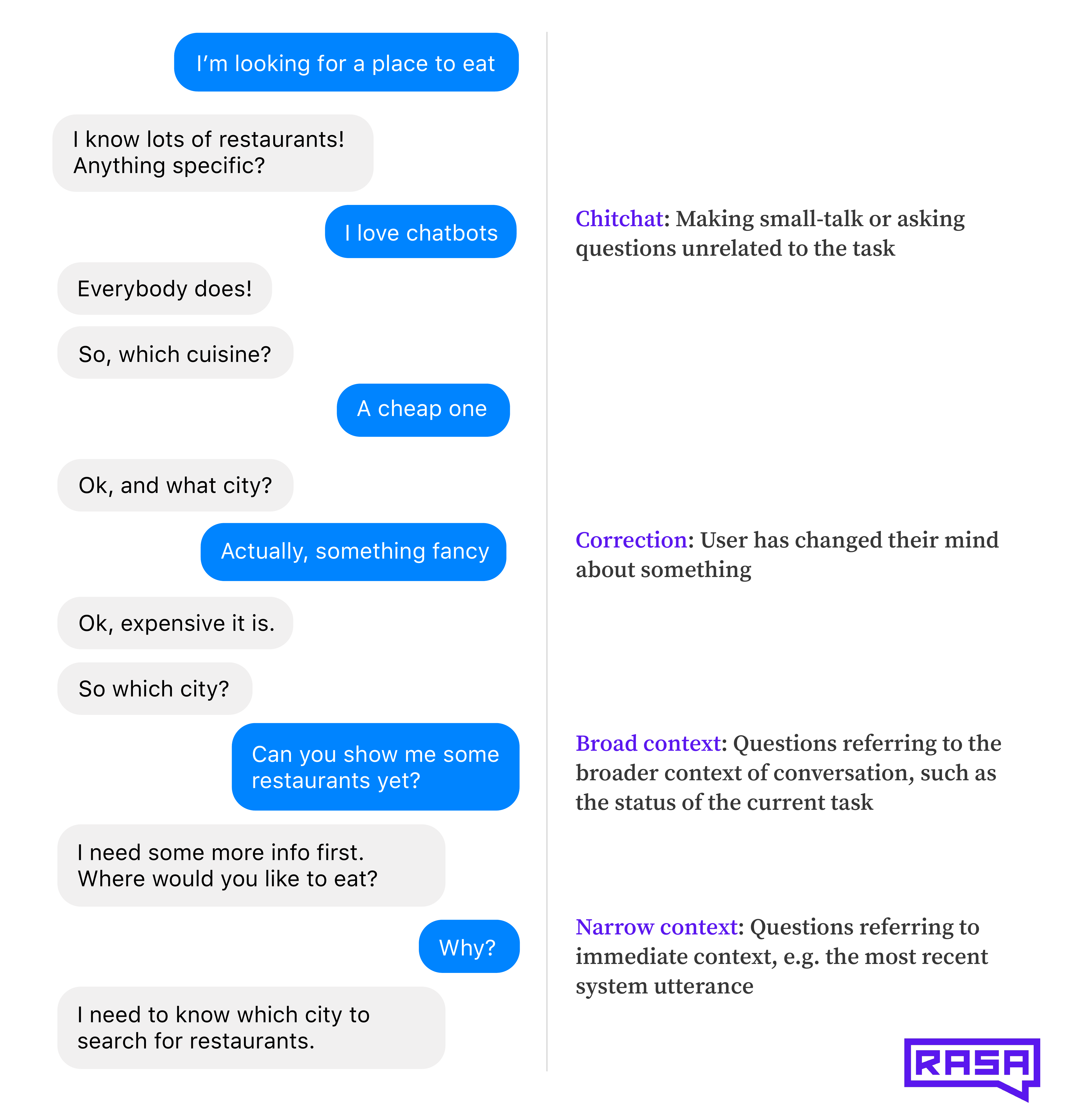}
\caption{An illustration of uncooperative dialogue, where the user does not provide the requested information. This shows each of the four deviation types we consider in our experiments.}
\label{fig:deviations}
\end{figure}

Figure~\ref{fig:deviations} exemplifies some common ways users deviate from cooperative dialogue, drawn from our experience of building dialogue systems. Each type of  deviation has to be handled differently, and the correct reaction is context-dependent. In each case the system should respond to the user's interjection, and then attempt to return to the task. When a user interjects with a question referring to the immediate dialogue context, e.g. \emph{why do you need to know that?}, the system must choose the appropriate response that explains why this information is required. In the case of an interjection referring to the broad context, e.g. \emph{can you show me some restaurants yet?}, the system must choose the appropriate response dependent on whether all required slots are filled. When the user provides a correction, e.g. \emph{actually, something fancy}, the system should acknowledge this update. Our model for {\tt chitchat} (non task-oriented conversation) is the simplest case: the dialogue policy should respond with an appropriate small-talk utterance, and then re-state its previous question. Specific chitchat responses are selected by an auxiliary system and are immaterial for our experiments. 

We follow the approach of Williams et al.~\citep{williams2017hybrid}, where a developer uses their domain knowledge to write an initial set of dialogues to be used for supervised learning. In our experiments in Section~\ref{sec:experiments}, we test the ability of a dialogue system exposed to uncooperative dialogues in one domain to transfer this knowledge to a new domain. 

\section{Related Work}
\paragraph{Modular versus end-to-end approaches}

There is a significant body of literature on constructing modular dialogue systems comprising language understanding, state tracking, action selection, and language generation~\citep{WILLIAMS2007393, young2013pomdp, levin2000stochastic, singh2002optimizing, williams2008best, hori2009statistical, lee2009example, griol2008statistical}. One successful approach has been to model dialogues as a Partially Observable Markov Decision Process (POMDP)~\citep{young2013pomdp, henderson2014second}. A POMDP maintains a distribution over possible states, where a state encodes the user's goal, the dialogue history, and the intent of the user's most recent message. POMDPs are usually trained in a Reinforcement Learning (RL) setting by engaging with a simulated user. Implementing a high-fidelity simulated user and reward function is a daunting task for non-expert application developers, so we emphasize supervised learning (SL) over RL, although our policy could also be refined using RL if desired.

Hybrid Code Networks (HCN)~\citep{williams2017hybrid} are a recently proposed mechanism for interfacing a machine learning-based dialogue manager with functions written by a developer. In our work, we similarly cast dialogue management as a problem of selecting the correct action to execute from a closed list. Actions can be as simple as sending a message to the user, but in general are arbitrary functions to execute, such as calling an API or querying a database. Note that we do not use statistical natural language generation, system utterances are simple templates. HCNs cannot directly be applied to our tasks, since we allow the dialogue manager to execute multiple actions in succession, and the policy must predict when it should `listen' (i.e. wait for the next user input). Though not identical, our LSTM classifier discussed in Section~\ref{sec:experiments} serves as a proxy for the HCN approach. HCNs are trained using a combination of SL and RL, where the initial SL data is manually created by the developer, and further SL data is collected on-line as users interact with the system. This builds on the hybrid SL/RL work of Henderson et al.~\citep{henderson2005hybrid}. We take a similar approach by bootstrapping with hand-crafted data, and then investigate how well dialogue policies are able to generalize from these initial examples to interactions with a simulated user. 

Recently multiple authors have proposed training end-to-end task-oriented dialogue managers~\citep{bordes2016learning, wen2016network}. For our purposes, a model is `end-to-end' if there is no requirement for domain-specific code and the policy can be trained on unannotated example dialogues. While this offers a promising path towards building adaptable cross-domain dialogue agents, both~\citep{bordes2016learning} and~\citep{wen2016network} point out that domain-specific slot definitions are still required to fully solve their tasks. 
Our approach is not end-to-end, but is trained on dialogues annotated with system actions and dialogue acts (i.e. the output of a Natural Language Understanding (NLU) system). An example dialogue is shown in figure~\ref{fig:attention}.

\paragraph{Domain Adaptation}

Prior work on cross-domain transfer learning in dialogue has focused on dialogue state tracking. State tracking involves inferring a belief state of the user goal from the sequence of user and system utterances, for example by using an RNN trained on ngram features~\citep{henderson2014word}. Gašic et al.~\citep{gavsic2014incremental} created a Gaussian Process-based POMDP system which can learn to integrate new slots within an existing domain. Mrkšić and collaborators~\citep{mrkvsic2015multi} used a recurrent neural network (RNN) approach to create a multi-domain dialogue state tracker. They showed that models trained on combined domains perform better than those trained on a single domain. Similar results were found by Williams~\citep{williams2013multi} using a maximum entropy approach. The primary challenge of dialogue state tracking is handling the uncertainties in spoken language understanding, and is a different task from the one studied in this work. 
Though NLU is an active area of research, the focus of this work is the dialogue policy, so we do not consider NLU errors.

\paragraph{Classification versus Ranking}

Unlike in~\citep{williams2017hybrid} and~\citep{wen2016network}, we do not use a classifier to select a system action. Instead, we jointly train embeddings for the dialogue state and each of the system actions, and use a similarity function between the state and the actions to rank each candidate. Other authors have applied ranking algorithms to dialogue management. Williams used a ranking approach based on decision trees for state tracking~\citep{williams2014web}. Bordes et al use embedding-based ranking as a `supervised embeddings' baseline in~\citep{bordes2016learning}, but in an end-to-end fashion where user and system utterances are embedded as bags of words. Banchs~\citep{banchs2012iris} applied an embed-and-rank approach to open-domain, non task-oriented dialogue, using a damping heuristic to combine the embedding vectors of the current utterance with the dialogue history. Unlike in~\citep{banchs2012iris, bordes2016learning}, we embed both the dialogue state and actions at a higher level of abstraction. User inputs are represented as dialogue acts, and system actions as a bag of features (see Section~\ref{sec:redp}). Unlike previous work using embeddings, our architecture contains a recurrent component to build up a representation of dialogue state. 

\paragraph{Tasks}
Our goal is to transfer learned behaviors from one slot-filling task (domain) to another. Some of the subtasks we consider have been considered within a single domain. For example, multiple authors have studied the interleaving of task and non-task conversation in dialogue systems~\citep{Papaioannou:2017, yu2017learning}, which is similar to our task of recovering gracefully from unprompted `chitchat' from users. In addition, our task of `handling corrections' is analogous to task 2 (updating API calls) defined for the restaurant domain in~\citep{bordes2016learning}.

\section{Recurrent Embedding Dialogue Policy} \label{sec:redp}
We propose the Recurrent Embedding Dialogue Policy (REDP). The aim is to learn vector embeddings for dialogue states and system actions in a supervised setting. At inference time, the current state of the dialogue is compared to all possible system actions, and the one with the highest cosine similarity is selected. This scheme was inspired by the StarSpace algorithm~\citep{wu2017starspace}, but unlike in an information retrieval system, creating a representation of dialogue state as a simple bag of features is ineffective. Instead, we propose an architecture which attends over the history of the dialogue, learning which previous user utterances and system actions are important for deciding which action to take next. 

One advantage of this approach is that target labels can be represented as a bag of multiple features, allowing us to represent system actions as a composition of features. In general, the features describing a particular action can come from a number of sources, including the class hierarchy, the name of the action, and even features derived from the code itself (such as which functions are called). Whatever the source of the features, similar actions should have more features in common than dissimilar actions, and ideally reflect the structure of the domains. In our experiments we only derive features from the action name, either taking the whole name as a single feature, or splitting the name into tokens and representing it as a bag of words. For example, the actions \texttt{utter\_explain\_details\_hotel} and \texttt{utter\_explain\_details\_restaurant} tell the user what information they must provide to get a recommendation. In the bag-of-words featurization, these actions have 3 features in common, and differ by a single feature indicating the domain. While token-based featurization is sufficient to illustrate generalization performance of our architecture, we would in general propose generating these features programmatically as it is less error-prone. 

\begin{figure}
\includegraphics[width=\linewidth]{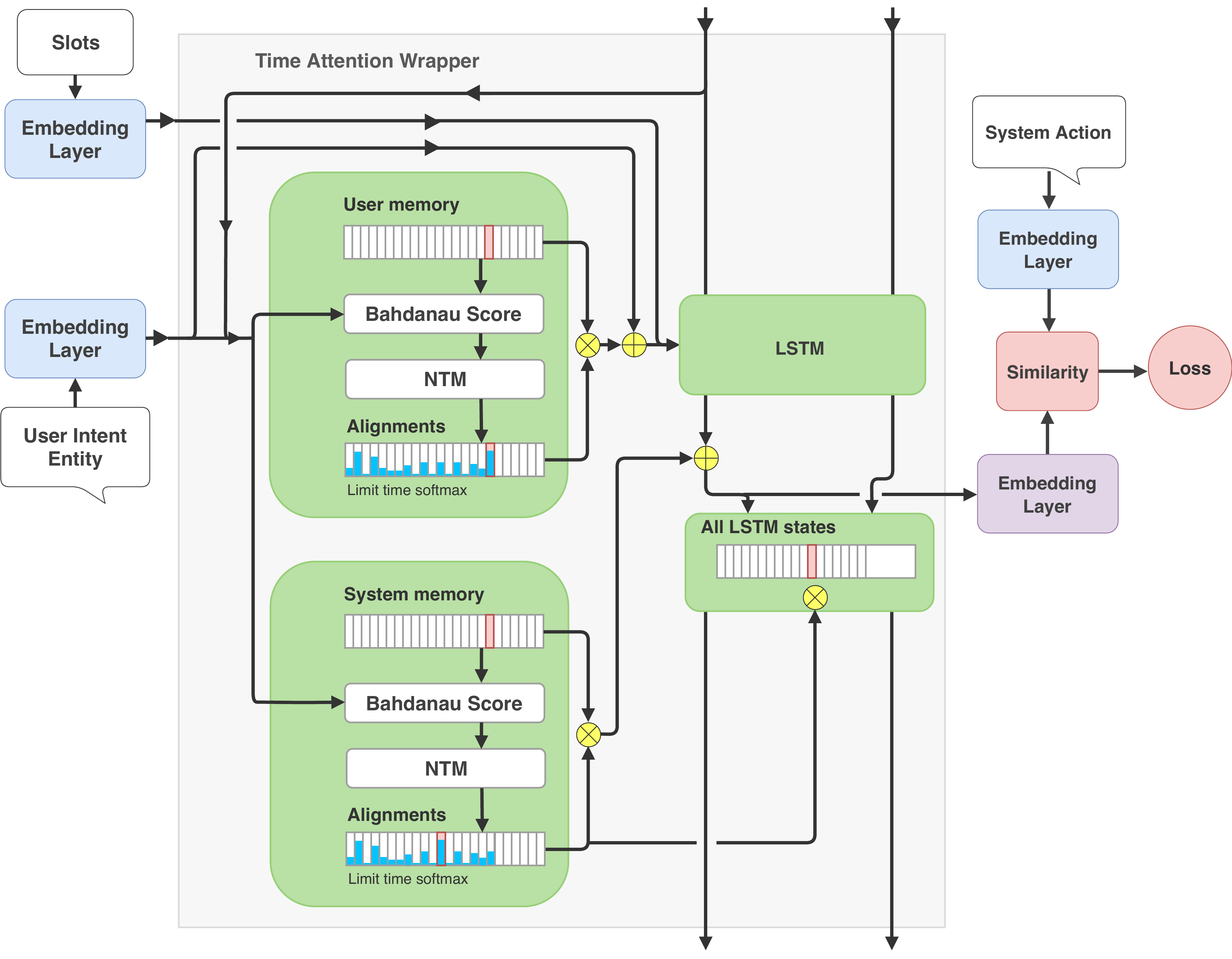}
\caption{The schematic representation of one time step of the  recurrent embedding dialogue policy. The signs $\otimes$ and $\oplus$ represent element-wise multiplication and summation respectively.}
\label{fig:embed_arch}
\end{figure}
One time step of our REDP is illustrated in Figure~\ref{fig:embed_arch}. It consists of several key steps.

\paragraph{Featurization}
The first step of the policy is to featurize user input, system actions and slots.

The labels for the user input are the intents and entities extracted by the natural language understanding system~\citep{bocklisch2017rasa}. The labels for the system actions are action names.
For our purposes, we use the tokens in the labels as features (e.g. \texttt{action\_search\_restaurant = \{action, search, restaurant\}}), but in general these features can be derived from any source.
A bag-of-words representations for the user and the system labels are then created using token counts inside each label.
The system actions are featurized separately.
The slots are featurized as binary vectors, indicating their presence or absence at each step of the dialogue. We use a simple slot tracking method, overwriting each slot with the most recently specified value. 

\paragraph{Embedding layers}
In the second step, embedding layers are applied to feature vectors to create embeddings for user input (intent and entities), system actions and slots.
The embedding layers are dense layers with separate weights for user input, slots, system actions and RNN output.

User input and the previous RNN output (the previous predicted system action embedding) are used to calculate an attention over a memory.

\paragraph{Attention}
An issue with standard attention mechanisms is that they attend to a fixed size memory, which in our case corresponds to both future and past time steps. While we can attend to both past and future actions during training, this is not possible at test time as future inputs are not known at the current time step. We therefore have to use an attention mechanism that only attends to past inputs and actions. In order to do this, the attention distribution is calculated over a truncated memory up to the current dialogue step.

Our attention architecture consists of a modified version of a Neural Turing Machine~\citep{graves2014neural}. The main difference is that since the memory is truncated at the current time step, the \emph{interpolation gate} cannot be applied to the attention probabilities. Therefore interpolation is applied to the Bahdanau scores~\citep{bahdanau2014neural} instead of the probabilities and only for previous time steps. Scores from the current time step are used without interpolation. The scores are calculated by using Bahdanau attention with normalization~\citep{salimans2016weight} as implemented in tensorflow~\citep{tensorflow2015-whitepaper}. To address memory by location, we perform 1-D convolution instead of circular convolution of attention probabilities with \emph{shift weighting} factors, which are calculated according as in Graves et. al~\citep{graves2014neural}. We use two separate attention mechanisms over user and system memories.

\paragraph{RNN}
As a cell for the RNN architecture, we use an LSTM cell with chrono initialization of forget and input biases introduced in~\citep{tallec2018can} and recurrent dropout on the hidden state~\citep{semeniuta2016recurrent}.

To create the input to the recurrent cell the embedded user input and user attention vector are summed, then concatenated with the embedded slots. The output of this cell is fed to another embedding layer to create an embedding of the cell output for the current time step. The sum of this embedded cell output and system attention vector is used as the dialogue state embedding.

At each time step, all previous LSTM states are multiplied element-wise with the binarized system attention probabilities. This way the architecture can learn to return to one of its previous LSTM states, skipping an arbitrary number of steps in between.

\paragraph{Similarity}
For each time step, a similarity is constructed between the dialogue level embedding $a$ and the embedding of the target system action $b_{+}$. We calculate the loss by negative sampling of embeddings of incorrect system actions $b_{-}$, ensuring that the similarity to the target label is high, and that the maximum similarity with any incorrect actions is low. 

The loss function is defined as:
\begin{equation}
\label{eq:loss}
L_t = \max\biggl(\mu_{+} - \textrm{sim}(a,b_{+}), 0\biggr) + \max\biggl(\mu_{-} + \max_{b_{-}}\bigl(\textrm{sim}(a,b_{-})\bigr), 0\biggr).
\end{equation}
This implies that the similarity for a positive system action would be maximized up to the value of $\mu_{+}$, and the maximum similarity over negative system actions would be minimized to $-\mu_{-}$.
The global loss is an average of all loss functions $L_t$ from all time steps.

\section{Experiments}
\label{sec:experiments}
We test the ability of LSTM and our REDP architectures to generalize each of the patterns shown in figure~\ref{fig:deviations} from one slot filling task to another. For illustration purposes, we consider the domains of restaurant and hotel recommendations. For both tasks, the system requires a price range, location, and number of people. To make a hotel recommendation, the system additionally needs to know the start and end date, and for a restaurant recommendation a cuisine type is also required.

\subsection{Data Generation}

We generated a number of dialogues by hand for use in supervised learning, as a developer would do to create an initial version of their dialogue system.
We created two kinds of dialogues: cooperative dialogues $d_{c}$ (of which there are 11), where the user may provide some or all of their preferences immediately, but otherwise responds to the system's questions without any deviations, and uncooperative dialogues $d_{u}^{\mathrm{init}}$, where exactly one deviation occurs (of one of the four types shown in figure~\ref{fig:deviations}).
We then trained a bootstrapped dialogue policy on these dialogues $d_{c} \cup d_{u}^{\mathrm{init}}$. 

We then created a simple, stochastic simulated user to interact with the bootstrapped policy. The user may cooperate and provide the requested information, or may interject with one of the deviation types. At each turn, the user action is sampled from a uniform distribution over these 5 response types. 
We generated 120 dialogues in the hotel domain and after removing  duplicates we had 108 unique dialogues. Since the bootstrapped policy is imperfect, we manually fixed the errors in these generated dialogues. We split these into a test set $d_{u}^{h,\mathrm{test}}$ of 30 dialogues and a training set $d_{u}^{h,\mathrm{train}}$ of 78 dialogues. For the restaurant domain we generated 50 uncooperative dialogues $d_{u}^{r}$ and 8 cooperative dialogues $d_{c}^{r}$ which we include in or exclude from the training data in order to measure the transfer learning abilities of the dialogue policies. The generated dialogues have a minimum of 12, maximum of 48, and mean of 29.4 actions to predict.

\subsection{Evaluation}
We compare two different policies in our experiments: an LSTM classifier which serves as a proxy for HCN, and our REDP policy as described in Section~\ref{sec:redp}. The input vector for the LSTM classifier is a binary vector encoding the user intent and entities, the slots, and the previous system action. In the LSTM(bin) baseline, the previous action is encoded as a one-hot vector. We also evaluate the LSTM(lt) policy, where the previous action is encoded as a bag of tokens. This was to ensure that the LSTM baseline had access to the same information as the REDP policy, namely the shared features between actions in different domains. 

We perform experiments on two different datasets,
combining the initial cooperative dialogues with uncooperative dialogues from the simulated user. The first dataset covers only the hotel domain, $d_{1} = d_{c}^{h} \cup d_{u}^{h,\mathrm{train}}$ and the second also includes dialogues in the restaurant domain $d_{2} = d_{1} \cup d_{c}^{r} \cup d_{u}^{r}$. We incrementally increase the fraction of $d_{u}^{h,\mathrm{train}}$, and evaluate on $d_{u}^{h,\mathrm{test}}$. We measure accuracy as the number of dialogues in which every action is predicted correctly, averaged over 5 independent runs.

All dialogue policies were implemented using the Rasa Core~\citep{bocklisch2017rasa} library~\url{https://github.com/RasaHQ/rasa_core}.
The code and data for our experiments are open source and available on the web \url{https://github.com/RasaHQ/conversational-ai-workshop-18}.
All hyperparameters are documented together with the code.

\begin{figure}
\begin{subfigure}{.48\textwidth}
  \includegraphics[width=\linewidth]{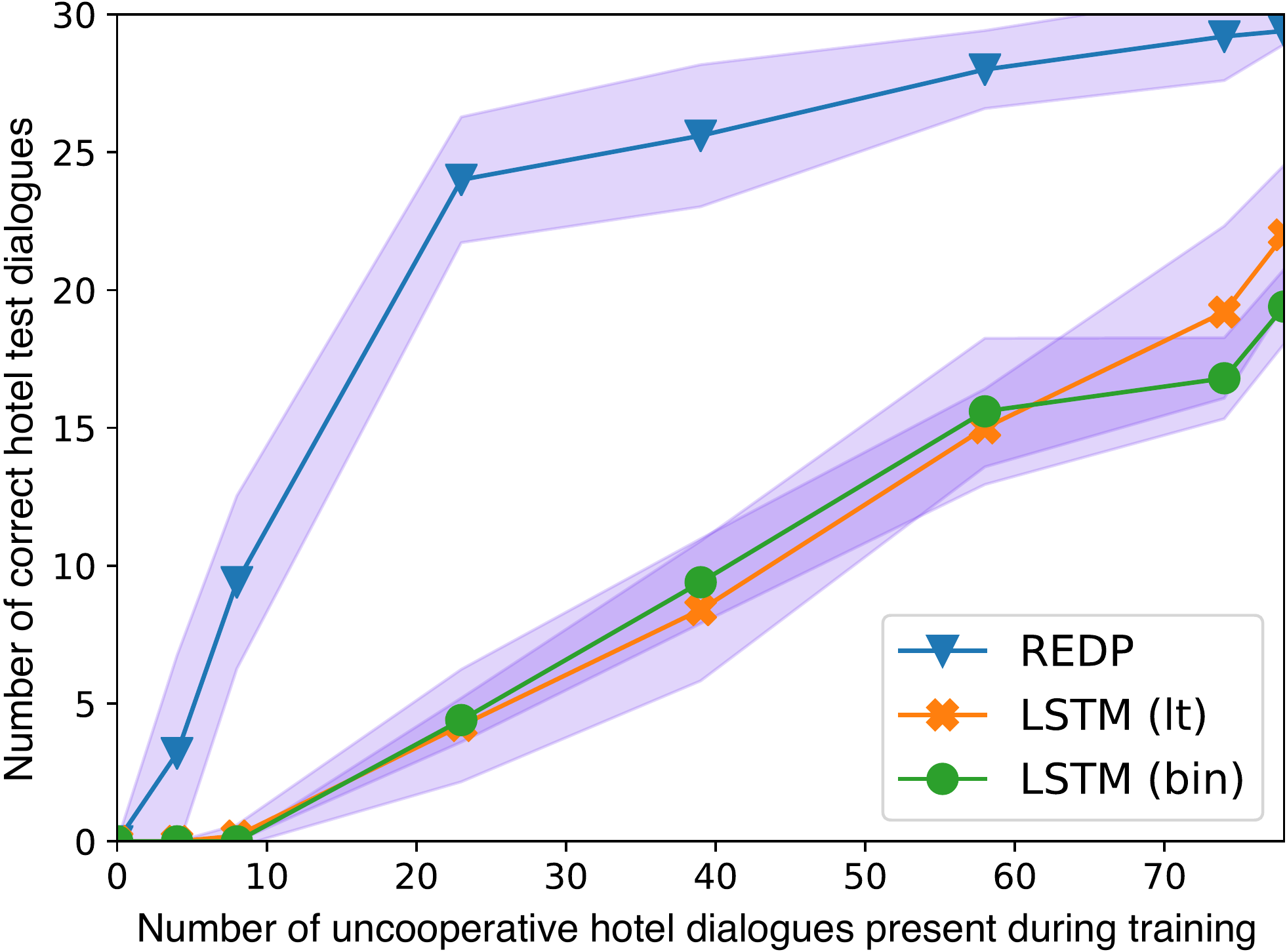}
  \caption{training data without restaurant dialogues ($d_1$)}
  \label{fig:generalise_noresto}
\end{subfigure} \ \ \ %
\begin{subfigure}{.48\textwidth}
  \includegraphics[width=\linewidth]{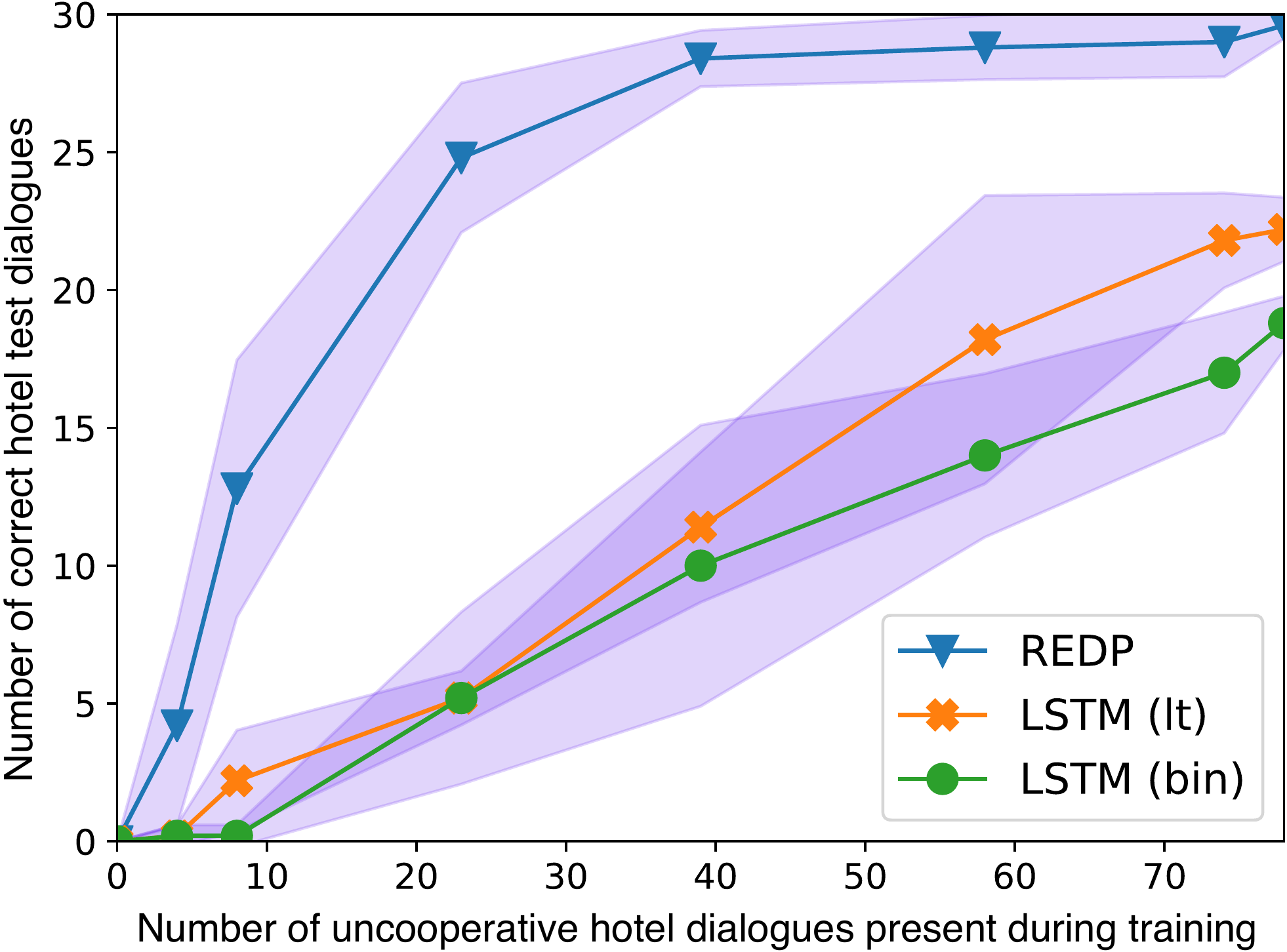}
  \caption{training data with restaurant dialogues ($d_2$)}
  \label{fig:generalise_resto}
\end{subfigure}

\caption{Generalization performance of the policies on two different datasets. On the horizontal axis is the number of uncooperative hotel dialogs $d_{u}^{h,\mathrm{train}}$ present in the training data. The vertical axis shows the number of uncooperative hotel dialogues in $d_{u}^{h,\mathrm{test}}$ in which every action is correctly predicted. Lines indicate the mean and the shaded area indicates one standard deviation.}
\label{fig:generalise}
\end{figure}

\begin{figure}
\begin{subfigure}{.32\textwidth}
\includegraphics[width=\linewidth]{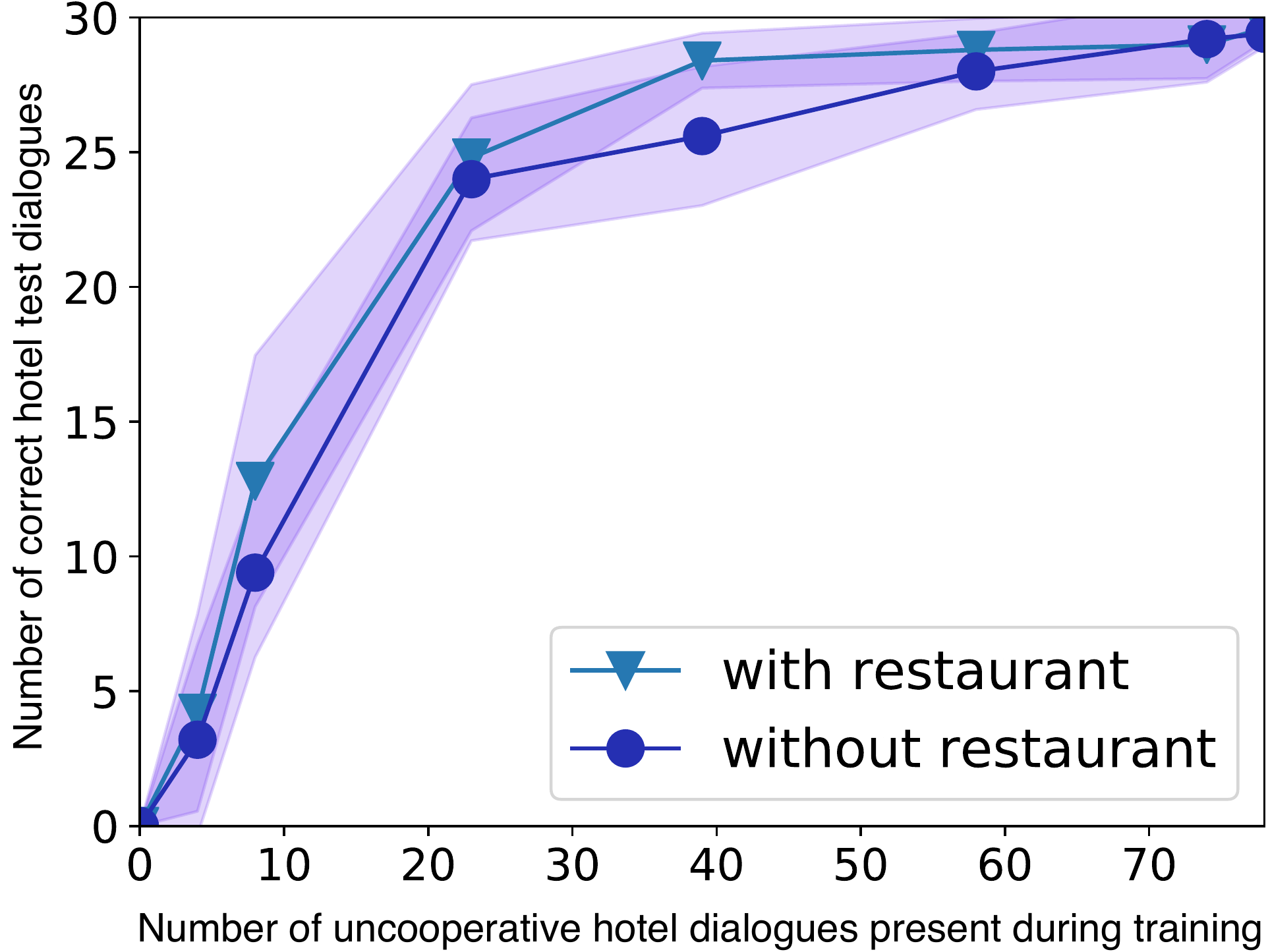}
  \caption{REDP}
  \label{fig:generalise_redp}
\end{subfigure} \ \ %
\begin{subfigure}{.32\textwidth}
\includegraphics[width=\linewidth]{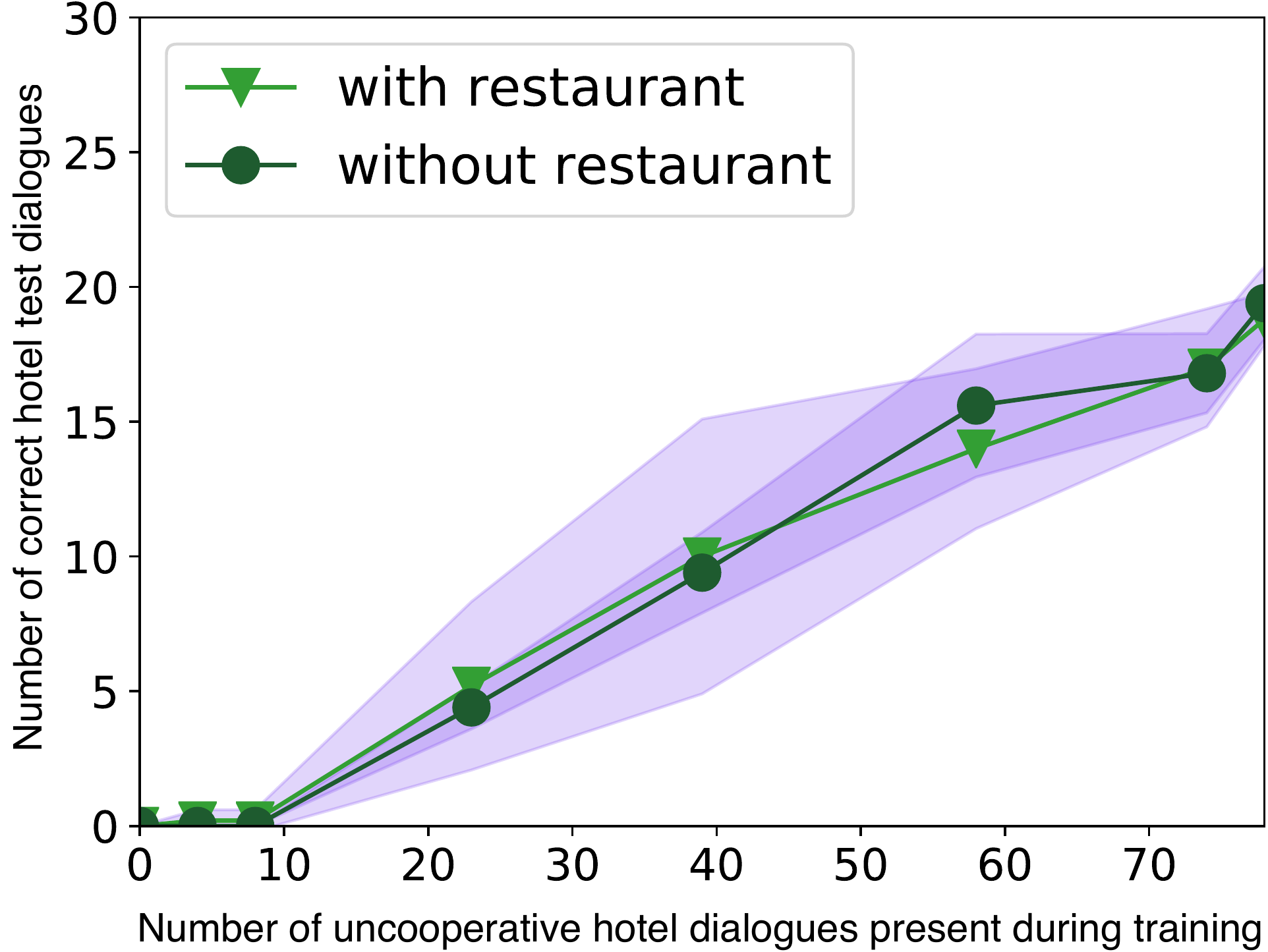}
  \caption{LSTM (bin)}
  \label{fig:generalise_keras_bin}
\end{subfigure} \ \ %
\begin{subfigure}{.32\textwidth}
\includegraphics[width=\linewidth]{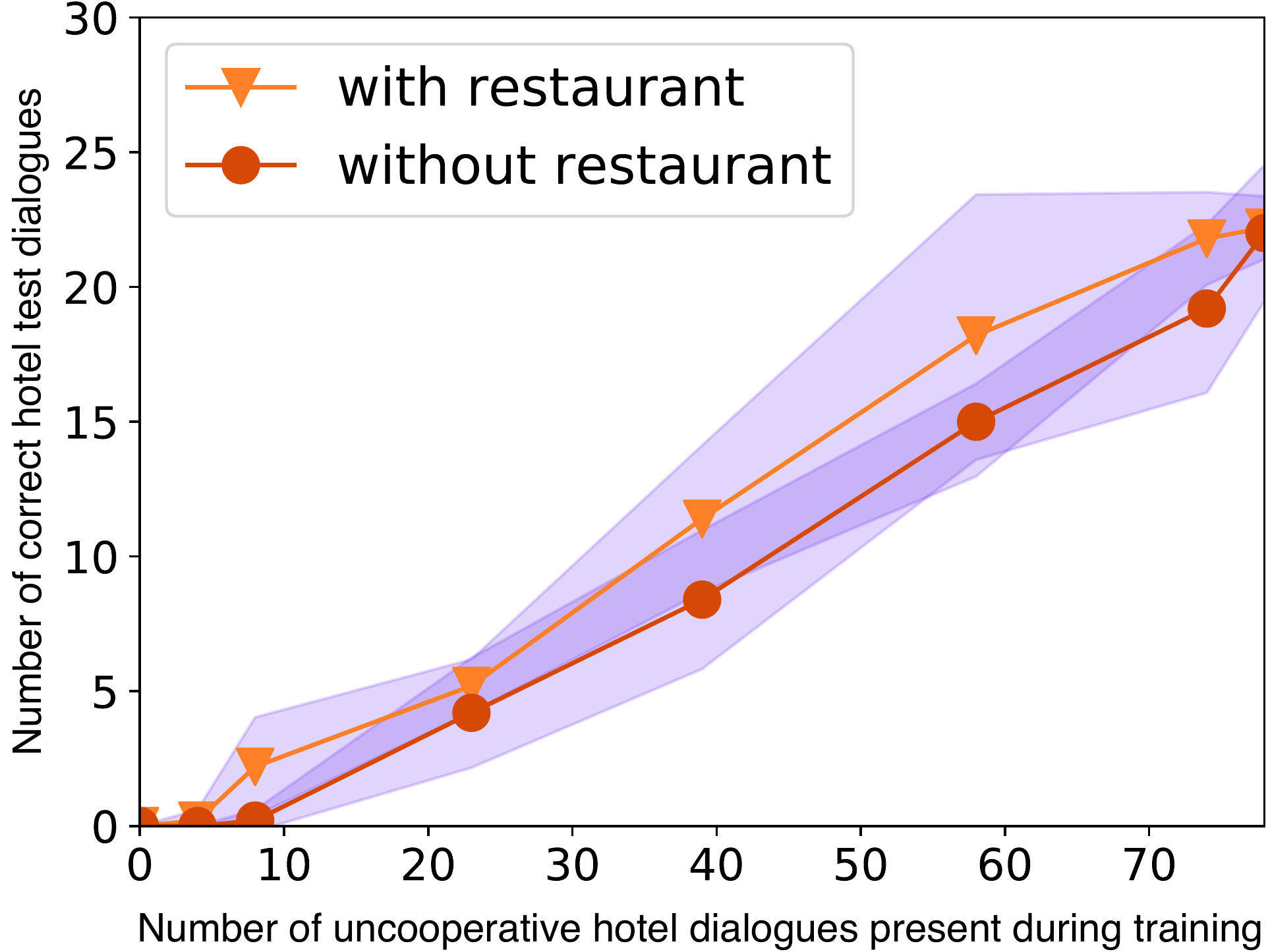}
  \caption{LSTM (lt)}
  \label{fig:generalise_keras}
\end{subfigure}
\caption{A direct comparison of the individual policies' performances on the training data with and without restaurant data, demonstrating transfer learning capabilities. The data displayed is the same as in Figure~\ref{fig:generalise}.}
\label{fig:w_wo_resto}
\end{figure}

\begin{figure}
\centering
\includegraphics[width=0.6\textwidth]{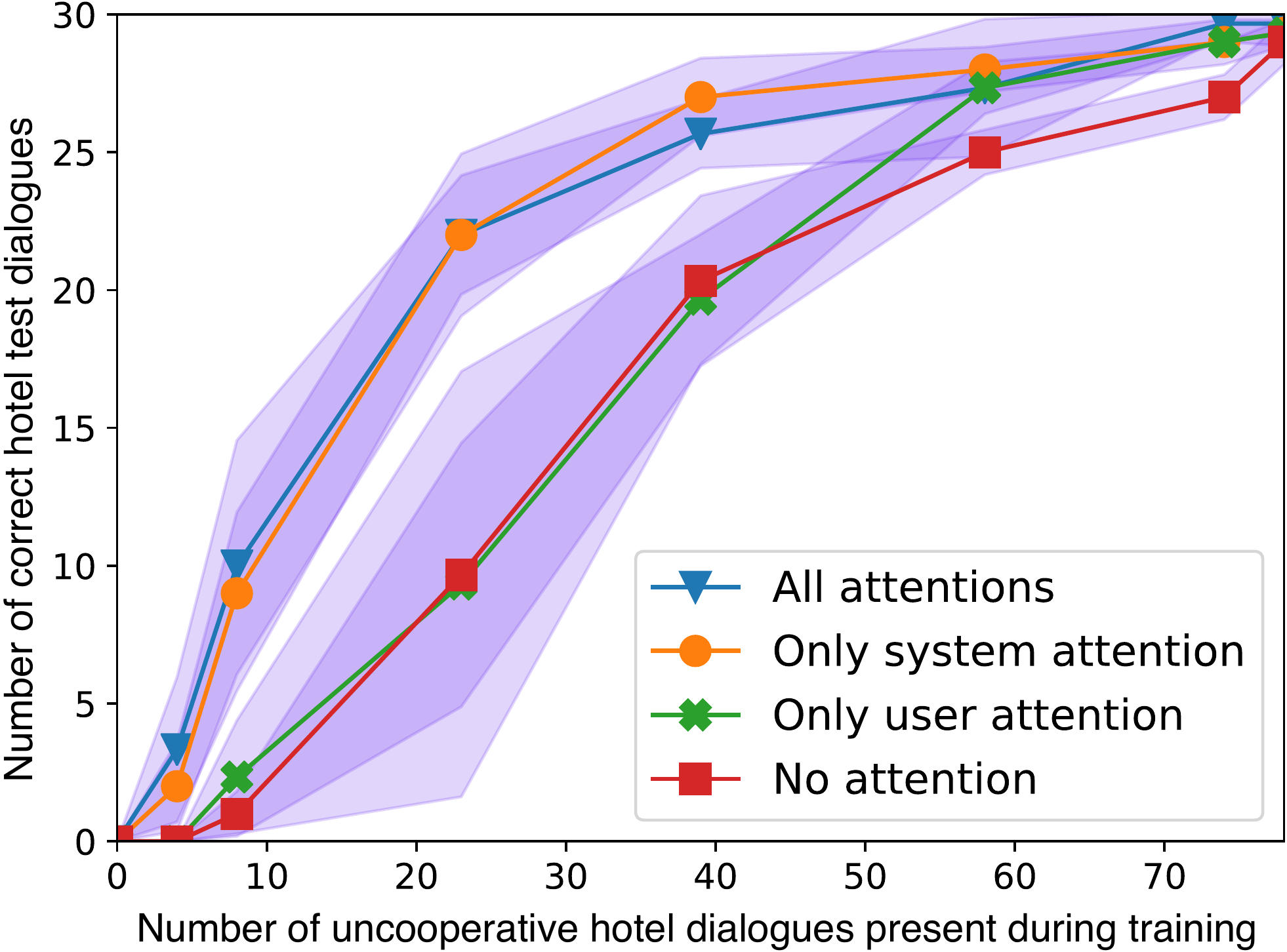}
\caption{Ablation study of REDP trained on $d_{1}$. For the tasks presented in this paper, attention over previous system actions is the key mechanism that enables better generalization on small amounts of training data.}
\label{fig:ablation}
\end{figure}

\begin{figure}
\includegraphics[width=\linewidth]{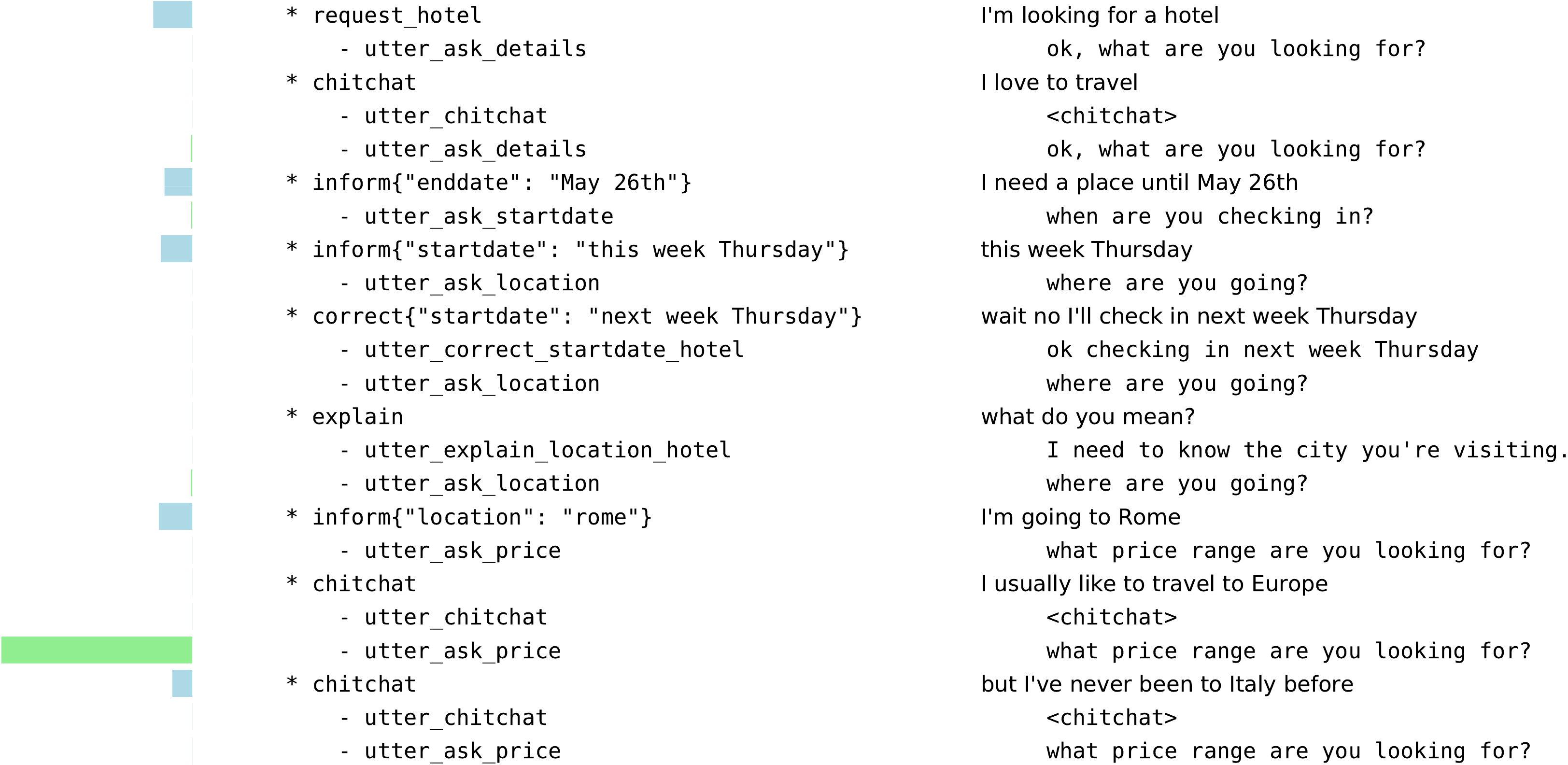}
\caption{Illustration of the role of the attention mechanism in handling uncooperative dialogue. On the left is a histogram of the alignments over system (solid bar) and user (hatched bar) memory when predicting the final action. In the center is a dialogue from the test set in Rasa Core format. On the right are utterances to help the reader follow the conversation. {\tt chitchat} responses must be selected separately and are not shown. The alignments illustrate that the policy can learn to ignore earlier uncooperative dialogue and complete the task.}
\label{fig:attention}
\end{figure}

\subsection{Results and Discussion}
Figures~\ref{fig:generalise_noresto} and~\ref{fig:generalise_resto} show the performance of each of the dialogue policies on the held-out test dialogues in the hotel domain. In Figure~\ref{fig:generalise_noresto} policies are trained on $d_1$, which only includes hotel dialogues, whereas in Figure~\ref{fig:generalise_resto} the policies were trained on $d_2$, which includes dialogues in the restaurant domain. 
REDP clearly outperforms the LSTM baselines on this task in both cases.

Figures~\ref{fig:generalise_redp}~-~\ref{fig:generalise_keras} show the same data as ~\ref{fig:generalise_noresto} and~\ref{fig:generalise_resto}, but grouped to illustrate the effect of transfer learning on each policy. While there is some evidence that REDP and LSTM(lt) benefit from the presence of restaurant dialogues, the evidence is not conclusive. However, the learning curve of REDP in Figure~\ref{fig:generalise_noresto} is already very steep,  leaving less room for improvement from transfer learning compared with the LSTM baseline. The most common error in the test set is an incorrect response to a narrow context question (see figure~\ref{fig:deviations}) where the main difficulty is identifying which slot the user is referring to. Tests with real users are required to determine the benefits of cross-task learning.

An ablation study of REDP~(shown in Figure~\ref{fig:ablation}) shows that the key mechanism for better generalization on the task presented in this paper is attention over previous system actions. Results were averaged over 3 runs. To demonstrate this, we illustrate in Figure~\ref{fig:attention} the influence of the attention mechanisms on recovering from uncooperative dialogue: by selectively attending to the dialogue history, the policy is able to recover from a sequence of uncooperative user utterances and return to the task to be completed. Note that even without any attention, REDP reaches a final test accuracy of 100\% which neither of the LSTM baselines achieve.

We also tested both policies on bAbI task 5 (conducting full dialogues) from~\citep{bordes2016learning}. We converted the dialogues into the Rasa Core format by automatically labeling the generated user and system utterances. Both policies achieve $100\%$ accuracy on the test set. This is to be expected since the bAbI dialogue task  was designed to challenge end-to-end systems and our dialogue policies use explicit slot tracking.

\section{Conclusions and Future Work}
In this paper, we have introduced the Recurrent Embedding Dialogue Policy, an approach which learns to embed dialogue states and dialogue system actions into the same vector space, and uses shared information in tasks to benefit from transfer learning across domains. We demonstrated that REDP can learn reusable dialogue patterns, and can leverage its attention mechanism to recover gracefully from uncooperative user behavior. We plan to study the REDP framework in more depth, applying it to more domains and tests with real users, and to study the properties of the learned embeddings.

\subsubsection*{Acknowledgments}
We wish to thank Peter Welinder and Verena Rieser for helpful comments on the manuscript. We also acknowledge interesting discussions with Tom Bocklisch and Joey Faulkner, who also helped us implement this work in Rasa Core. Special thanks to Elise Boyd for supporting us with the illustrations. We would like to thank the Rasa community for feedback and support.

\clearpage

\end{document}